\documentclass[10pt,twocolumn,letterpaper]{article}
\usepackage{cvpr}
\usepackage[pagebackref=true,breaklinks=true,letterpaper=true,colorlinks,bookmarks=false]{hyperref}
\usepackage{times}
\usepackage{epsfig}
\usepackage{graphicx}
\usepackage{amsmath}
\usepackage{amssymb}
\usepackage{booktabs}

\usepackage{amsmath,amsfonts,bm}

\def\eqref#1{equation~\ref{#1}}

\def\1{\bm{1}}

\def\eps{{\epsilon}}

\def\valpha{{\bm{\alpha}}}
\def\vphi{{\bm{\phi}}}

\def\vw{{\bm{w}}}
\def\vx{{\bm{x}}}

\DeclareMathAlphabet{\mathsfit}{\encodingdefault}{\sfdefault}{m}{sl}
\SetMathAlphabet{\mathsfit}{bold}{\encodingdefault}{\sfdefault}{bx}{n}

\DeclareMathOperator*{\argmin}{arg\,min}

\usepackage{algorithm}
\usepackage{algpseudocode}
\usepackage{array,multirow,graphicx}
\newcommand{\task}{\mathcal{T}}

\newtheorem{theor}{Theorem}

\newcommand{\eat}[1]{}

\title{Meta-Learning with Differentiable Convex Optimization}

\author{Kwonjoon Lee$^{2}$\qquad Subhransu Maji$^{1,3}$\qquad Avinash Ravichandran$^{1}$\qquad Stefano Soatto$^{1,4}$  \\
$^1$Amazon Web Services\qquad 
$^2$UC San Diego\qquad 
$^3$UMass Amherst\qquad 
$^4$UCLA\\
\texttt{kwl042@ucsd.edu}\qquad \texttt{\{smmaji,ravinash,soattos\}@amazon.com} \\
}

\cvprfinalcopy 

\def\cvprPaperID{3073} 

\ifcvprfinal\fi

\begin{document}

\maketitle
\thispagestyle{empty}
\begin{abstract}
Many meta-learning approaches for few-shot learning
rely on simple base learners such as nearest-neighbor classifiers.
However, even in the few-shot regime, discriminatively trained linear
predictors can offer better generalization.
We propose to use these predictors as base learners to learn
representations for few-shot learning and show they offer better
tradeoffs between feature size and performance across a range of
few-shot recognition benchmarks.
Our objective is to learn feature embeddings that generalize well
under a linear classification rule for novel categories. 
To efficiently solve the objective, we exploit two properties of
linear classifiers: implicit differentiation of the optimality
conditions of the convex problem and the dual formulation of the
optimization problem.
This allows us to use high-dimensional embeddings with improved
generalization at a modest increase in computational overhead.
Our approach, named MetaOptNet, achieves state-of-the-art performance on
miniImageNet, tieredImageNet, CIFAR-FS, and FC100 few-shot learning
benchmarks. Our code is available online\footnote{\url{https://github.com/kjunelee/MetaOptNet}}.
\end{abstract}
\vspace{-3mm}
\section{Introduction}
\label{sec:intro}
The ability to learn from a few examples is a hallmark of human intelligence, 
yet it remains a challenge for modern machine learning systems.
This problem has received significant attention from the machine
learning community recently where few-shot learning is cast as a
meta-learning problem (\eg,~\cite{meta-lstm,MAML,matching-net,proto-net}). 
The goal is to minimize generalization
error across a distribution of tasks with few training examples.
Typically, these approaches are composed of an \emph{embedding model}
that maps the input domain into a feature space and a \emph{base learner}
that maps the feature space to task variables.
The meta-learning objective is to learn an embedding model such that
the base learner generalizes well across tasks.

While many choices for base learners exist, nearest-neighbor
classifiers and their variants (\eg,~\cite{proto-net,matching-net}) are popular as the
classification rule
is simple and the approach scales well in the low-data regime.
However, discriminatively trained linear classifiers often outperform
nearest-neighbor classifiers (\eg,~\cite{Caruana:2008:EES:1390156.1390169,exemplar-svm})
in the low-data regime as they can exploit
the negative examples which are often more abundant to learn better
class boundaries. 
Moreover, they can effectively use high dimensional
feature embeddings as model capacity can be controlled by appropriate
regularization such as weight sparsity or norm.

\begin{figure*}[!htp]
\vspace{-19mm} 
\begin{center}
\begin{tabular} {c}
\includegraphics[width=1.0\textwidth]{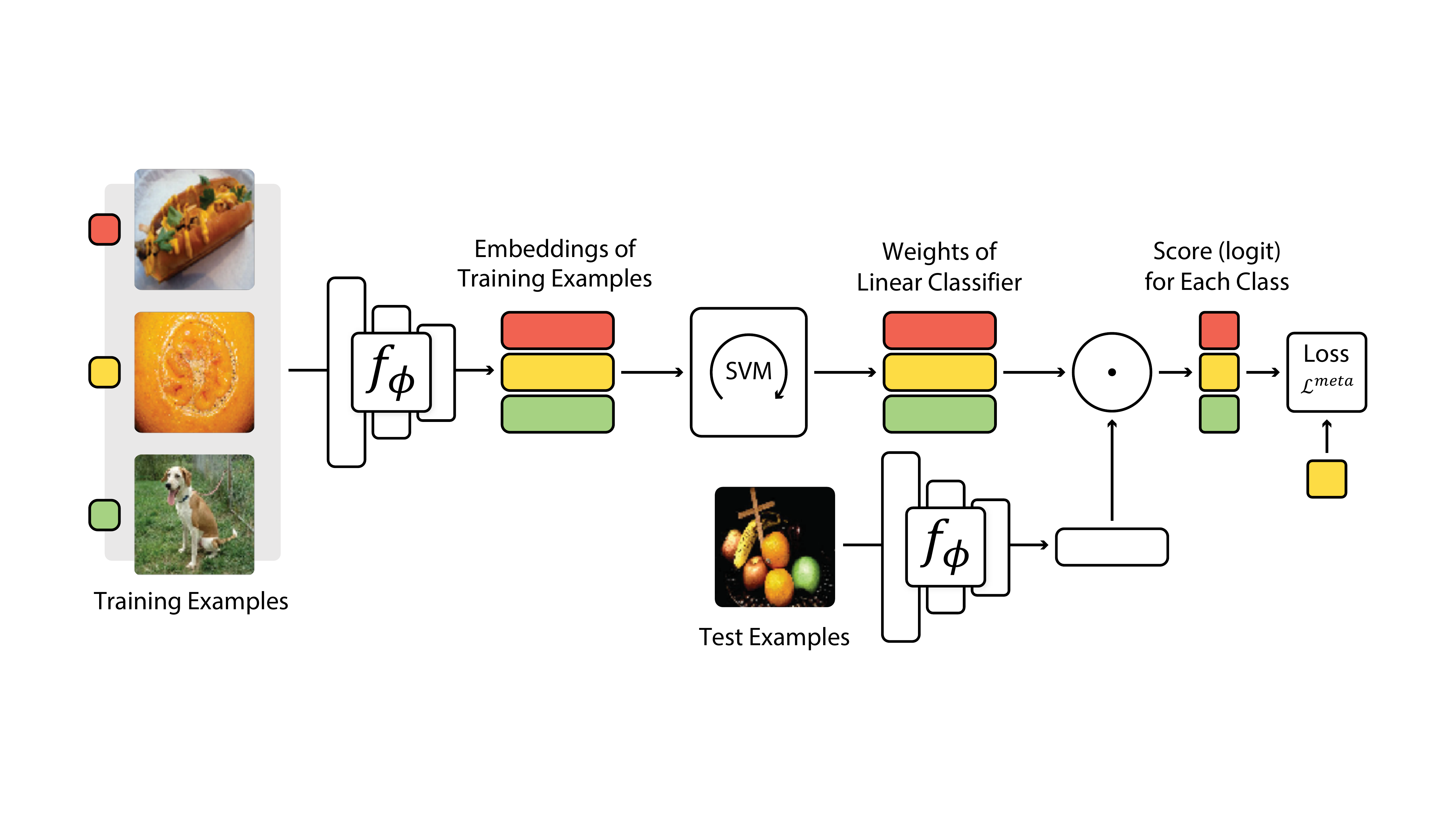} 
\end{tabular}
\end{center}
\vspace{-22mm}
\caption{\textbf{Overview of our approach.} Schematic illustration of our method MetaOptNet on an 1-shot 3-way classification task. The meta-training objective is to learn the parameters $\phi$ of a feature embedding model $f_\phi$ that generalizes well across tasks when used with regularized linear classifiers (\eg, SVMs). A task is a tuple of a few-shot training set and a test set (see Section \ref{sec:approach} for details).}
\label{fig:MetaOptNet_pipeline}
\end{figure*}

Hence, in this paper, we investigate linear classifiers as the base
learner for a meta-learning based approach for few-shot learning.
The approach is illustrated in Figure~\ref{fig:MetaOptNet_pipeline}
where a linear support vector machine (SVM) is used to learn a classifier given a set of
labeled training examples and the generalization error is
computed on a novel set of examples from the same task.
The key challenge is computational since the meta-learning
objective of minimizing the generalization error across tasks
requires training a linear classifier in the inner loop of
optimization (see Section~\ref{sec:approach}). 
However, the objective of linear models is convex and can be solved
efficiently.
We observe that two additional properties arising from the convex
nature that allows efficient meta-learning: implicit differentiation
of the optimization \cite{implicit1, gould2016differentiating} and the
low-rank nature of the classifier in the few-shot setting. 
The first property allows the use of off-the-shelf convex optimizers
to estimate the optima and implicitly differentiate the optimality or Karush-Kuhn-Tucker
(KKT) conditions to train embedding model.
The second property means that the number of optimization variables in the \emph{dual
formation} is far smaller than the feature dimension for few-shot learning.

To this end, we have incorporated a \emph{differentiable quadratic programming} (QP) solver~\cite{amos2017optnet} which allows end-to-end learning of the embedding model with various linear classifiers, \eg, multiclass support vector machines (SVMs) \cite{CS} or linear regression, for few-shot classification tasks.
Making use of these properties, we show that our method is practical and offers substantial gains over nearest neighbor classifiers at a modest increase in computational costs (see Table \ref{tab:imagenet-tradeoff}). 
Our method achieves state-of-the-art performance on 5-way 1-shot and 5-shot classification for popular few-shot benchmarks including miniImageNet \cite{matching-net, meta-lstm}, tieredImageNet \cite{ren2018metalearning}, CIFAR-FS \cite{R2D2}, and FC100 \cite{TADAM}.

\section{Related Work}
\label{sec:related}
Meta-learning studies what aspects of the learner
(commonly referred to as \textit{bias} or \textit{prior}) effect generalization across a distribution of
tasks~\cite{schmidhuber:1987:srl, Thrun1998, Vilalta2002}.
Meta-learning approaches for few-shot learning can be broadly
categorized these approaches into three groups. 
\emph{Gradient-based methods}~\cite{meta-lstm,MAML} use gradient descent
to adapt the embedding model parameters (\eg, all layers of a deep network) given
training examples. 
\emph{Nearest-neighbor methods} \cite{matching-net, proto-net} learn
a distance-based prediction rule over the embeddings. For
example, prototypical networks~\cite{proto-net} represent each class
by the mean embedding of the examples, and the classification rule is
based on the distance to the nearest class mean.
Another example is matching networks~\cite{matching-net} that learns a
kernel
density estimate of the class densities using the embeddings over
training data (the model can also be interpreted
as a form of \emph{attention} over training examples).
\emph{Model-based methods} \cite{SNAIL, pmlr-v80-munkhdalai18a} learn a
parameterized predictor to estimate model parameters, \eg, a recurrent network that predicts
parameters analogous to a few steps of gradient descent in parameter space. 
While gradient-based methods are general, they are prone to
overfitting as the embedding dimension grows~\cite{SNAIL,LEO}.
Nearest-neighbor approaches offer simplicity and scale well in the
few-shot setting. 
However, nearest-neighbor methods have no mechanisms for feature
selection and are not very robust to noisy features. 

Our work is related to techniques for \emph{backpropagation} though optimization procedures. 
Domke \cite{pmlr-v22-domke12} presented a generic method based on
\emph{unrolling} gradient-descent for a fixed number of steps and
automatic differentiation to compute gradients.
However, the trace of the optimizer (\ie, the intermediate values) needs to be stored in order to
compute the gradients which can be prohibitive for large problems.
The storage overhead issue was considered in more detail by Maclaurin
\etal~\cite{maclaurin2015gradient} where they studied low precision
representations of the optimization trace of deep networks.
If the {\rm argmin} of the optimization can be found analytically, such as in unconstrained quadratic minimization problems, then it is also possible to compute the gradients analytically.
This has been applied for learning in low-level vision
problems~\cite{tappen2007learning,schmidt2014shrinkage}.
A concurrent and closely related work \cite{R2D2} uses this idea to
learn few-shot models using ridge-regression base learners which have
closed-form solutions.
We refer readers to Gould \etal~\cite{gould2016differentiating} which
provides an excellent survey of techniques for differentiating {\rm
  argmin} and {\rm argmax} problems.

Our approach advocates the use of linear classifiers which can be
formulated as convex learning problems.
In particular, the objective is a \emph{quadratic program} (QP) which can be efficiently
solved to obtain its \emph{global optima} using gradient-based techniques.
Moreover, the solution to convex problems can be characterized by their
Karush-Kuhn-Tucker (KKT) conditions which allow us to
\emph{backpropagate} through the learner using the \emph{implicit function theorem}~\cite{krantz2012implicit}.
Specifically, we use the formulation of Amos and
Kolter~\cite{amos2017optnet} which provides efficient GPU routines for computing solutions to QPs and
their gradients. 
While they applied this framework to learn representations for
constraint satisfaction problems, it is also well-suited for few-shot learning as the
problem sizes that arise are typically small.

While our experiments focus on linear classifiers with hinge loss
and $\ell_2$ regularization, our framework can be used with other
loss functions and non-linear kernels.
For example, the ridge regression learner used in~\cite{R2D2} can
be implemented within our framework allowing a direct comparison.

\section{Meta-learning with Convex Base Learners}
\label{sec:approach}
We first derive the meta-learning framework for
few-shot learning following prior work (\eg, ~\cite{proto-net,meta-lstm,MAML}) and then discuss how
convex base learners, such as linear SVMs, can be incorporated.

\subsection{Problem formulation}
Given the training set $\mathcal{D}^{train}=\{(\vx_t,y_t)\}_{t=1}^T$,
the goal of the base learner $\mathcal{A}$ is to estimate parameters
$\theta$ of the predictor $y=f(\vx;\theta)$ so that it generalizes
well to the unseen test set
$\mathcal{D}^{test}=\{(\vx_t,y_t)\}_{t=1}^Q$.
It is often assumed that the training and test set are sampled from
the same distribution and the domain is mapped to a feature space
using an embedding model $f_\phi$ parameterized by $\phi$. For
optimization-based learners, the parameters are obtained by minimizing
the empirical loss over training data along with a regularization that
encourages simpler models. 
This can be written as:
\begin{equation}
\theta=\mathcal{A}(\mathcal{D}^{train}; \phi)=\argmin_{\theta}\mathcal{L}^{base}(\mathcal{D}^{train};\theta,\phi) + \mathcal{R}(\theta)
\end{equation}
where $\mathcal{L}^{base}$ is a loss function, such as the negative log-likelihood of labels, and $\mathcal{R}(\theta)$ is a regularization term. Regularization plays an important role in generalization when training data
is limited.

Meta-learning approaches for few-shot learning aim to minimize the
generalization error across a distribution of tasks sampled from a
task distribution. Concretely, this can be thought of as learning over a collection of
tasks:
$\mathcal{T}=\{(\mathcal{D}_i^{train},\mathcal{D}_i^{test})\}_{i=1}^I$, 
often referred to as a \textit{meta-training} set.
The tuple $(\mathcal{D}_i^{train},\mathcal{D}_i^{test})$ describes a
training and a test dataset, or a task.
The objective is to learn an embedding model $\phi$ that minimizes
generalization (or test) error across tasks given a base learner
${\cal A}$.
Formally, the learning objective is:
\begin{equation}
\min_{\phi} \mathbb{E}_{{\mathcal T}}  
  \left[ \mathcal{L}^{meta}(\mathcal{D}^{test}; \theta, \phi)
    ,\text{where } \theta=\mathcal{A}(\mathcal{D}^{train};
    \phi)\right].
\label{eq:meta-objective}
\end{equation}

Figure~\ref{fig:MetaOptNet_pipeline} illustrates the training and
testing for a single task.
Once the embedding model $f_\phi$ is learned, its generalization is
estimated on a set of \emph{held-out} tasks (often referred to as a
\textit{meta-test} set)
$\mathcal{S}=\{(\mathcal{D}_j^{train},\mathcal{D}_j^{test})\}_{j=1}^J$
computed as:
\begin{equation}
\mathbb{E}_{{\mathcal S}}  
  \left[ \mathcal{L}^{meta}(\mathcal{D}^{test}; \theta, \phi) ,\text{where } \theta=\mathcal{A}(\mathcal{D}^{train}; \phi)\right].
\label{eq:meta-test}
\end{equation}
Following prior work~\cite{meta-lstm,MAML}, we call the stages of estimating the expectation in
Equation~\ref{eq:meta-objective} and \ref{eq:meta-test} as meta-training and meta-testing respectively.
During meta-training, we keep an additional held-out
\textit{meta-validation} set to choose the hyperparameters of the
meta-learner and pick the best embedding model.

\subsection{Episodic sampling of tasks} 
Standard few-shot learning benchmarks such as miniImageNet
\cite{meta-lstm} evaluate models in $K$-way, $N$-shot classification
tasks.
Here $K$ denotes the number of classes, and $N$ denotes the number of training examples per class.
Few-shot learning techniques are evaluated for small values of $N$, typically $N\in\{1, 5\}$.
In practice, these datasets do not explicitly contain tuples
$(\mathcal{D}_i^{train},\mathcal{D}_i^{test})$, but each task for
meta-learning is constructed ``on the fly" during the meta-training
stage, commonly described as an episode.

For example, in prior work \cite{matching-net, meta-lstm}, a task (or episode) $\task_i=(\mathcal{D}_i^{train},\mathcal{D}_i^{test})$ is sampled as follows.
The overall set of categories is $C^{train}$. 
For each episode, categories $C_i$ containing $K$ categories from the
$C^{train}$ are first sampled (with replacement); then training
(support) set $\mathcal{D}_i^{train}=\{(\vx_n, y_n) \mid
n=1,\ldots,N\times K, y_n\in C_i\}$ consisting of $N$ images per
category is sampled; and finally, the test (query) set
$\mathcal{D}_i^{test}=\{(\vx_n, y_n) \mid n=1,\ldots,Q\times K, y_n\in
C_i\}$ consisting of $Q$ images per category is sampled.

We emphasize that we need to sample without replacement, \ie,
$\mathcal{D}_i^{train}\cap \mathcal{D}_i^{test}=\O$, to optimize the
generalization error. In the same manner, meta-validation set and
meta-test set are constructed on the fly from $C^{val}$ and
$C^{test}$, respectively. In order to measure the embedding model's
generalization to \emph{unseen categories}, $C^{train}$, $C^{val}$, and
$C^{test}$ are chosen to be mutually disjoint.

\subsection{Convex base learners}
The choice of the base learner ${\cal A}$ has a significant impact on
Equation~\ref{eq:meta-objective}.
The base learner that computes $\theta = \mathcal{A}(D^{train}; \phi)$ has to be
efficient since the expectation has to be computed over a
distribution of tasks. 
Moreover, to estimate parameters $\phi$ of the embedding model the
gradients of the task test error 
 $\mathcal{L}^{meta}(\mathcal{D}^{test}; \theta, \phi)$ with respect
to $\phi$ have to be efficiently computed.
This has motivated simple base learners such as nearest class
mean~\cite{proto-net} for which the parameters of the base learner $\theta$ are easy to
compute and the objective is differentiable.

We consider base learners based on multi-class linear classifiers
(\eg, support vector machines (SVMs)~\cite{CS, WW}, logistic
regression, and ridge regression) where the base-learner's objective is
\emph{convex}.
For example, a $K$-class linear SVM can be written as
$\theta=\{\vw_k\}_{k=1}^K$. The Crammer and Singer~\cite{CS}
formulation of the multi-class SVM is:
\begin{equation}
\begin{split}
&\theta=\mathcal{A}(\mathcal{D}^{train};\vphi)=\argmin_{\{ \vw_k \}}\min_{\{ \xi_i \}} \frac{1}{2}\sum_k{||\vw_k||_2^2+C\sum_n \xi_n} \\
&\text{subject\;to} \\
&\vw_{y_n}\cdot f_\vphi(\vx_n)-\vw_{k}\cdot f_\vphi (\vx_n) \ge 1-\delta_{y_n,k}-\xi_n,~\forall n, k
\end{split}
\label{eq:CS-objective}
\end{equation}
where $\mathcal{D}^{train}=\{(\vx_n, y_n)\}$, $C$ is the regularization parameter and $\delta_{\cdot, \cdot}$ is the Kronecker delta function.

\paragraph{Gradients of the SVM objective.}
From Figure \ref{fig:MetaOptNet_pipeline}, we see that in order to
make our system end-to-end trainable, we require that the solution of
the SVM solver should be differentiable with respect to its input,
\ie, we should be able to compute $\{\frac{\partial \theta}{\partial
  f_\vphi(\vx_n)}\}_{n=1}^{N\times K}$. 
The objective of SVM is convex and has a unique optimum.
This allows for the use of implicit function theorem
(\eg,~\cite{krantz2012implicit,dontchev2009implicit,implicit1}) on the
optimality (KKT) conditions to obtain the necessary gradients. 
For the sake of completeness, we derive the form of the theorem for
convex optimization problems as stated in~\cite{implicit1}. 
Consider the following convex optimization problem:
\begin{equation}
\begin{aligned}
& \text{minimize} & & f_0(\theta, z) \\
&  \text{subject to} && f(\theta, z) \preceq 0 \\
&&& h(\theta,z)=0.
\end{aligned}
\label{eq:convex_objective}
\end{equation}
where the vector $\theta \in \mathbb{R}^d$ is the optimization variable of the problem, the vector $z \in \mathbb{R}^e$ is the input parameter of the optimization problem, which is $\{f_\phi(\vx_n)\}$ in our case. We can optimize the objective by solving for the saddle point $(\tilde{\theta},\tilde\lambda,\tilde\nu)$ of the following Lagrangian:
\begin{equation}
L(\theta,\lambda,\nu,z) = f_0(\theta,z) + \lambda^T f(\theta,z) + \nu^T h(\theta,z).
\end{equation}
In other words, we can obtain the optimum of the objective function by solving $g(\tilde\theta,\tilde\lambda,\tilde\nu,z)=0$ where
\begin{equation}
g(\theta,\lambda,\nu,z) = \begin{bmatrix}
\nabla_\theta L({\theta},\lambda,\nu,z) \\
\textbf{diag}(\lambda) f(\theta,{z}) \\
h(\theta,{z})
\end{bmatrix}.
\end{equation}

Given a function $f(x): \mathbb{R}^n \rightarrow \mathbb{R}^m$, denote
$D_x f(x)$ as its Jacobian $\in \mathbb{R}^{m \times n}$.

\begin{theor} 
(From Barratt~\cite{implicit1})
\label{th:implicit_theorem}
Suppose $g(\tilde{\theta},\tilde\lambda,\tilde\nu,z)=0$.
Then, when all derivatives exist,
\begin{equation}
D_z \tilde{\theta} = -D_\theta g(\tilde{\theta},\tilde\lambda,\tilde\nu,z)^{-1} D_z g(\tilde{\theta},\tilde\lambda,\tilde\nu,z).
\label{eq:implicit_theorem}
\end{equation}
\end{theor}

This result is obtained by applying the implicit function theorem to the KKT conditions.
Thus, once we compute the optimal solution $\tilde{\theta}$, we can obtain a closed-form expression for the gradient of $\tilde{\theta}$ with respect to the input data.
This obviates the need for backpropagating through the entire optimization trajectory since the solution does not depend on the trajectory or initialization due to its uniqueness.
This also saves memory, an advantage that convex problems have over generic optimization problems.

\paragraph{Time complexity.}
The forward pass (\ie, computation of Equation~\ref{eq:CS-objective}) using our approach requires the solution to the QP solver whose complexity scales as $O(d^3)$ where $d$ is the number of optimization variables. 
This time is dominated by factorizing the KKT matrix required for primal-dual interior point method. 
Backward pass requires the solution to
Equation~\ref{eq:implicit_theorem} in
Theorem~\ref{th:implicit_theorem}, whose complexity is $O(d^2)$ given
the factorization already computed in the forward pass. Both forward
pass and backward pass can be expensive when the dimension of embedding $f_\phi$ is large.

\paragraph{Dual formulation.}
The dual formulation of the objective in
Equation~\ref{eq:CS-objective} allows us to address
the poor dependence on the embedding dimension and can be written as follows.
Let \begin{equation}
  \label{eq:vw-def}
  \vw_{k}(\valpha^k)=\sum_n \alpha_n^k f_\phi(\vx_n) \quad \forall \ k.
\end{equation} 
We can instead optimize in the dual space:
\begin{equation}
\label{eq:dual-CS}
\begin{split}
\max_{\{ \valpha^k \}}&\;\Big[-\frac{1}{2}\sum_{k}||\vw_{k}(\valpha^k)||_2^2+\sum_{n} \alpha_n^{y_n} \Big] \\
\text{subject\;to}& \\
&\alpha_n^{y_n} \leq C, \quad \alpha_n^{k} \leq 0 \quad \forall k \neq y_n, \\
&\sum_{k}\alpha_n^k= 0 \quad \forall n.
\end{split}
\end{equation}

This results in a quadratic program (QP) over the dual variables $\{ \valpha^k \}_{k=1}^K$.
We note that the size of the optimization variable is the number of training examples times the number of classes. 
This is often \emph{much smaller} than the size of the feature dimension for few-shot learning. 
We solve the dual QP of Equation \ref{eq:dual-CS} using \cite{amos2017optnet} which implements a differentiable GPU-based QP solver. 
In practice (as seen in Table \ref{tab:imagenet-tradeoff}) the time taken by the QP solver is comparable to the time taken to compute features using the ResNet-12 architectures so the overall speed per iteration is not significantly different from those based on simple base learners such as nearest class prototype (mean) used in Prototypical Networks~\cite{proto-net}.

Concurrent to our work, Bertinetto \etal~\cite{R2D2} employed ridge
regression as the base learner which has a closed-form solution.
Although ridge regression may not be best suited for classification
problems, their work showed that training models by minimizing squared
error with respect to one-hot labels works well in practice.
The resulting optimization for ridge regression is also a QP and can
be implemented within our framework as:
\begin{equation}
\label{eq:dual-RR}
    \max_{\{ \valpha^k \}}\Big[-\frac{1}{2}\sum_{k}||\vw_{k}(\valpha^k)||_2^2-\frac{\lambda}{2}\sum_{k}||\valpha^k||_2^2+\sum_{n} \alpha_n^{y_n} \Big]
\end{equation}
where $\vw_{k}$ is defined as Equation \ref{eq:vw-def}.
A comparison of linear SVM and ridge regression in
Section~\ref{experiements} shows a slight advantage of the linear SVM formation.

\subsection{Meta-learning objective}
To measure the performance of the model we evaluate the negative log-likelihood of the test data sampled from the same task.
Hence, we can re-express the meta-learning objective of Equation~\ref{eq:meta-objective} as:
\begin{equation}
\begin{aligned}
  &\mathcal{L}^{meta}(\mathcal{D}^{test}; \theta,\phi,\gamma)= \\
& \sum_{(\vx,y)\in\mathcal{D}^{test}}[-\gamma\vw_y\cdot f_{\phi}(\vx)+\log\sum_k \exp(\gamma\vw_k\cdot f_{\phi}(\vx))]
\end{aligned}
\label{eq:meta-objective_2}
\end{equation}
where $\theta=\mathcal{A}(\mathcal{D}^{train}; \phi)=\{\vw_k\}_{k=1}^K$ and $\gamma$ is a learnable scale parameter.
Prior work in few-shot learning \cite{TADAM, R2D2,
  gidaris2018dynamic} suggest that adjusting the prediction score by
a learnable scale parameter $\gamma$ provides better performance under
nearest class mean and ridge regression base learners. 

We empirically find that inserting $\gamma$ is beneficial for the meta-learning with SVM base learner as well.
While other choices of test loss, such as hinge loss, are possible,
log-likelihood worked the best in our experiments.

\section{Experiments}\label{experiements}
We first describe the network architecture and optimization details
used in our experiments (Section~\ref{sec:implementation}).
We then present results on standard few-shot classification benchmarks including
derivatives of ImageNet (Section~\ref{sec:exp:imagenet}) and CIFAR
(Section~\ref{sec:exp:cifar}), followed by a detailed analysis of the
impact of various base learners on accuracy and speed using the same
embedding network and training setup
(Section~\ref{sec:exp:ablation}-\ref{sec:exp:efficiency}).

\subsection{Implementation details}\label{sec:implementation}
\noindent\textbf{Meta-learning setup.} We use a ResNet-12 network following
\cite{TADAM, SNAIL} in our experiments.
Let \verb|Rk| denote a residual block that consists of three
$\{$3$\times$3 convolution with \verb|k| filters, batch normalization,
Leaky ReLU(0.1)$\}$; let \verb|MP| denote a 2$\times$2 max
pooling. We use DropBlock regularization~\cite{DropBlock}, a form of
structured Dropout. 
Let \verb|DB(k, b)| denote a DropBlock layer with keep\_rate\verb|=k|
and block\_size\verb|=b|. 
The network architecture for ImageNet derivatives is:
\verb|R64-MP-DB(0.9,1)-R160-MP-DB(0.9,1)-R320-|
\verb|MP-DB(0.9,5)-R640-MP-DB(0.9,5)|, 
while the network architecture used for CIFAR derivatives is: \verb|R64-MP-DB(0.9,1)-R160-MP-DB(0.9,1)-R320-|
\verb|MP-DB(0.9,2)-R640-MP-DB(0.9,2)|.
We do not apply a global average pooling after the last residual block.

As an optimizer, we use SGD with Nesterov momentum of 0.9 and
weight decay of 0.0005. 
Each mini-batch consists of 8 episodes. 
The model was meta-trained for 60 epochs, with each epoch
consisting of 1000 episodes. The learning rate was initially set to
0.1, and then changed to 0.006, 0.0012, and 0.00024 at
epochs 20, 40 and 50, respectively, following the practice of
\cite{gidaris2018dynamic}.

During meta-training, we adopt horizontal flip, random crop, and
color (brightness, contrast, and saturation) jitter
data augmentation as in \cite{gidaris2018dynamic, Act2Param}. 
For experiments on miniImageNet with ResNet-12, we use label smoothing
with $\eps=0.1$. Unlike \cite{proto-net} where they used higher
way classification for meta-training than meta-testing, we use a 5-way
classification in both stages following recent works
\cite{gidaris2018dynamic, TADAM}. 
Each class contains 6 test (query) samples during meta-training and 15 test
samples during meta-testing. Our meta-trained model was chosen based on
5-way 5-shot test accuracy on the meta-validation set.

\noindent\textbf{Meta-training shot.} 
For prototypical networks, we match the meta-training shot to
meta-testing shot following
the usual practice \cite{proto-net, gidaris2018dynamic}. For SVM and
ridge regression, we observe that keeping meta-training
shot higher than meta-testing shot leads to better test accuracies as shown in Figure \ref{fig:meta-training-shot}. 
Hence, during meta-training, we set training shot to 15 for miniImageNet with
ResNet-12; 5 for miniImageNet with 4-layer CNN (in Table
\ref{tab:imagenet-tradeoff}); 10 for tieredImageNet; 5 for CIFAR-FS;
and 15 for FC100.

\noindent\textbf{Base-learner setup.} For linear classifier
training, we use
the quadratic programming (QP) solver OptNet
\cite{amos2017optnet}. Regularization parameter $C$ of SVM was set to
$0.1$. Regularization parameter $\lambda$ of ridge regression was set
to $50.0$. For the nearest class mean (prototypical
networks), we use squared Euclidean distance normalized with respect
to the feature dimension.

\noindent\textbf{Early stopping.} 
Although we can run the optimizer until convergence, in practice we found
that running the QP solver for a fixed number of iterations (just
three) works well in practice.
Early stopping acts an additional regularizer and even leads to a
slightly better performance.

\subsection{Experiments on ImageNet derivatives}
\label{sec:exp:imagenet}
The \textbf{miniImageNet} dataset \cite{matching-net} is a standard
benchmark for few-shot image classification benchmark, consisting of
100 randomly chosen classes from ILSVRC-2012~\cite{imagenet}. 
These classes are randomly split into $64$, $16$ and $20$ classes for
meta-training, meta-validation, and meta-testing respectively.
Each class contains $600$ images of size
$84\times84$. 
Since the class splits were not released in the original
  publication~\cite{matching-net}, we use the commonly-used split
  proposed in~\cite{meta-lstm}.

The \textbf{tieredImageNet} benchmark \cite{ren2018metalearning} is a
larger subset of ILSVRC-2012 \cite{imagenet}, composed of 608 classes
grouped into 34 high-level categories. These are divided into $20$
categories for meta-training, $6$ categories for meta-validation, and $8$ categories for
meta-testing. This corresponds to $351$, $97$ and $160$ classes for
meta-training, meta-validation, and meta-testing respectively.
This dataset aims to minimize the semantic similarity between
the splits. All images are of size $84\times 84$.

\begin{table*}[tb]
\caption{\textbf{Comparison to prior work on miniImageNet and tieredImageNet.} Average few-shot classification accuracies (\%) with 95\% confidence intervals on miniImageNet and tieredImageNet meta-test splits. a-b-c-d denotes a 4-layer convolutional network with a, b, c, and d filters in each layer. \textsuperscript{$\ast$}Results from \cite{meta-lstm}. \textsuperscript{$\dagger$}Used the union of meta-training set and meta-validation set to meta-train the meta-learner. ``RR" stands for ridge regression.}
\label{tab:miniImagenet}
\begin{center}
\begin{small}
\begin{tabular}{@{}llc@{}cc@{}c@{}cc@{}}
\hline
\toprule
& & \phantom{a} & \multicolumn{2}{c}{\textbf{miniImageNet 5-way}} & \phantom{ab} & \multicolumn{2}{c}{\textbf{tieredImageNet 5-way}} \\
\cmidrule{4-5} \cmidrule{7-8}
\textbf{model} & \textbf{backbone} && \textbf{1-shot} & \textbf{5-shot} && \textbf{1-shot} & \textbf{5-shot}  \\
\hline
Meta-Learning LSTM\textsuperscript{$\ast$} \cite{meta-lstm} & 64-64-64-64 && 43.44 $\pm$ 0.77 & 60.60 $\pm$ 0.71 && - & - \\
Matching Networks\textsuperscript{$\ast$} \cite{matching-net} & 64-64-64-64 && 43.56 $\pm$ 0.84 & 55.31 $\pm$ 0.73 && - & - \\
MAML \cite{MAML} & 32-32-32-32 &&  48.70 $\pm$ 1.84 & 63.11 $\pm$ 0.92 && 51.67 $\pm$ 1.81 & 70.30 $\pm$ 1.75 \\
Prototypical Networks\textsuperscript{$\ast$}\textsuperscript{$\dagger$} \cite{proto-net} & 64-64-64-64 && 49.42 $\pm$ 0.78 & 68.20 $\pm$ 0.66 && 53.31 $\pm$ 0.89 & 72.69 $\pm$ 0.74 \\
Relation Networks\textsuperscript{$\ast$} \cite{sung2018learning} & 64-96-128-256 && 50.44 $\pm$ 0.82 & 65.32 $\pm$ 0.70 && 54.48 $\pm$ 0.93 & 71.32 $\pm$ 0.78 \\
R2D2 \cite{R2D2} & 96-192-384-512 && 51.2 $\pm$ 0.6 & 68.8 $\pm$ 0.1 && - & -\\
Transductive Prop Nets \cite{transductive-prop} & 64-64-64-64 && 55.51 $\pm$ 0.86 & 69.86 $\pm$ 0.65 && 59.91 $\pm$ 0.94 & 73.30 $\pm$ 0.75 \\
SNAIL \cite{SNAIL} & ResNet-12 && 55.71 $\pm$ 0.99 & 68.88 $\pm$ 0.92 && - & - \\
Dynamic Few-shot \cite{gidaris2018dynamic} & 64-64-128-128 && 56.20 $\pm$ 0.86 & 73.00 $\pm$ 0.64 && - & - \\
AdaResNet \cite{pmlr-v80-munkhdalai18a} & ResNet-12 && 56.88 $\pm$ 0.62 & 71.94 $\pm$ 0.57 && - & - \\
TADAM \cite{TADAM} & ResNet-12 && 58.50 $\pm$ 0.30 & 76.70 $\pm$ 0.30 && - & - \\
Activation to Parameter\textsuperscript{$\dagger$} \cite{Act2Param} & WRN-28-10 && 59.60 $\pm$ 0.41 & 73.74 $\pm$ 0.19 && - & - \\
LEO\textsuperscript{$\dagger$} \cite{LEO} & WRN-28-10 && 61.76 $\pm$ 0.08 & 77.59 $\pm$ 0.12 && \textbf{66.33 $\pm$ 0.05} & \textbf{81.44 $\pm$ 0.09} \\
MetaOptNet-RR (ours) & ResNet-12 && 61.41 $\pm$ 0.61 & 77.88 $\pm$ 0.46 && \textbf{65.36 $\pm$ 0.71} & \textbf{81.34 $\pm$ 0.52}\\
MetaOptNet-SVM (ours) & ResNet-12 && 62.64 $\pm$ 0.61 & 78.63 $\pm$ 0.46 && \textbf{65.99 $\pm$ 0.72} & \textbf{81.56 $\pm$ 0.53} \\
MetaOptNet-SVM-trainval (ours)\textsuperscript{$\dagger$} & ResNet-12 && \textbf{64.09 $\pm$ 0.62} & \textbf{80.00 $\pm$ 0.45} && \textbf{65.81 $\pm$ 0.74} & \textbf{81.75 $\pm$ 0.53} \\
\bottomrule
\hline
\end{tabular}
\vspace{-5mm}
\end{small}

\end{center}
\end{table*}

\noindent\textbf{Results}. Table~\ref{tab:miniImagenet} summarizes the
results on the 5-way miniImageNet and tieredImageNet. 
Our method achieves state-of-the-art performance on
5-way miniImageNet and tieredImageNet benchmarks. 
Note that LEO \cite{LEO} make use of encoder and relation
network in addition to the WRN-28-10 backbone network to produce
sample-dependent initialization of gradient descent. TADAM
\cite{TADAM} employs a task embedding network (TEN) block for each
convolutional layer -- which predicts element-wise scale and shift
vectors. 

We also note that \cite{LEO,Act2Param} pretrain the WRN-28-10 feature
extractor \cite{WRN} to jointly classify all 64 classes in
miniImageNet meta-training set; then freeze the network during the
meta-training. \cite{TADAM} make use of a similar strategy of using
standard classification: they co-train the feature embedding on
few-shot classification task (5-way) and standard classification task
(64-way). In contrast, our system is meta-trained end-to-end,
explicitly training the feature extractor to work well on few-shot
learning tasks with regularized linear classifiers. This strategy
allows us to clearly see the effect of meta-learning.
Our method is arguably simpler and achieves strong performance.

\subsection{Experiments on CIFAR derivatives}\label{sec:exp:cifar}
The \textbf{CIFAR-FS} dataset \cite{R2D2} is a recently proposed
few-shot image classification benchmark, consisting of all 100 classes
from CIFAR-100 \cite{cifar-100}. The classes are randomly split
into $64$, $16$ and $20$ for meta-training, meta-validation,
and meta-testing respectively. Each class contains $600$ images of
size $32\times32$.

The \textbf{FC100} dataset \cite{TADAM} is another dataset derived
from CIFAR-100 \cite{cifar-100}, containing 100 classes which are
grouped into 20 superclasses. These classes are partitioned into $60$
classes from $12$ superclasses for meta-training, $20$ classes from
$4$ superclasses for meta-validation, and $20$ classes from $4$
superclasses for meta-testing. The goal is to minimize semantic overlap
between classes similar to the goal of tieredImageNet.
Each class contains 600 images of size $32\times32$.

\noindent\textbf{Results}. Table~\ref{tab:CIFAR} summarizes the
results on the 5-way classification tasks where our method
MetaOptNet-SVM achieves the state-of-the-art performance.
On the harder FC100 dataset, the gap between various base learners is
more significant, which highlights the advantage of complex base
learners in the few-shot learning setting.

\begin{table*}[tb]
\caption{\textbf{Comparison to prior work on CIFAR-FS and FC100.} Average few-shot classification accuracies (\%) with 95\% confidence intervals on CIFAR-FS and FC100. a-b-c-d denotes a 4-layer convolutional network with a, b, c, and d filters in each layer. \textsuperscript{$\ast$}CIFAR-FS results from \cite{R2D2}. \textsuperscript{$\dagger$}FC100 result from \cite{TADAM}. \textsuperscript{$\mathparagraph$}Used the union of meta-training set and meta-validation set to meta-train the meta-learner. ``RR" stands for ridge regression.} 
\label{tab:CIFAR}
\begin{center}
\begin{small}
\begin{tabular}{@{}llc@{}cc@{}c@{}cc@{}}
\hline
\toprule
& & \phantom{a} & \multicolumn{2}{c}{\textbf{CIFAR-FS 5-way}} & \phantom{ab} & \multicolumn{2}{c}{\textbf{FC100 5-way}} \\
\cmidrule{4-5} \cmidrule{7-8}
\textbf{model} & \textbf{backbone} && \textbf{1-shot} & \textbf{5-shot} && \textbf{1-shot} & \textbf{5-shot}  \\
\hline
MAML\textsuperscript{$\ast$} \cite{MAML} & 32-32-32-32 &&  58.9 $\pm$ 1.9  & 71.5 $\pm$ 1.0  && - & - \\
Prototypical Networks\textsuperscript{$\ast$}\textsuperscript{$\dagger$} \cite{proto-net} & 64-64-64-64 && 55.5 $\pm$ 0.7 & 72.0 $\pm$ 0.6 && 35.3 $\pm$ 0.6 & 48.6 $\pm$ 0.6 \\
Relation Networks\textsuperscript{$\ast$} \cite{sung2018learning} & 64-96-128-256 && 55.0 $\pm$ 1.0 & 69.3 $\pm$ 0.8 && - & - \\
R2D2 \cite{R2D2} & 96-192-384-512 && 65.3 $\pm$ 0.2 & 79.4 $\pm$ 0.1 && - & -\\
TADAM \cite{TADAM} & ResNet-12 && - & - && 40.1 $\pm$ 0.4 & 56.1 $\pm$ 0.4\\
ProtoNets (our backbone) \cite{proto-net}& ResNet-12 && \textbf{72.2 $\pm$ 0.7} & 83.5 $\pm$ 0.5 && 37.5 $\pm$ 0.6 & 52.5 $\pm$ 0.6 \\
MetaOptNet-RR (ours) & ResNet-12 && \textbf{72.6 $\pm$ 0.7} & \textbf{84.3 $\pm$ 0.5} && 40.5 $\pm$ 0.6 & 55.3 $\pm$ 0.6\\
MetaOptNet-SVM (ours) & ResNet-12 && \textbf{72.0 $\pm$ 0.7} & \textbf{84.2 $\pm$ 0.5} && 41.1 $\pm$ 0.6 & 55.5 $\pm$ 0.6 \\
MetaOptNet-SVM-trainval (ours)\textsuperscript{$\mathparagraph$} & ResNet-12 && \textbf{72.8 $\pm$ 0.7} & \textbf{85.0 $\pm$ 0.5} && \textbf{47.2 $\pm$ 0.6} & \textbf{62.5 $\pm$ 0.6} \\
\bottomrule
\hline
\end{tabular}
\end{small}
\end{center}
\end{table*}

\begin{table*}[tb]
\caption{\textbf{Effect of the base learner and embedding network architecture.} Average few-shot classification accuracy (\%) and forward inference time (ms) per episode on miniImageNet and tieredImageNet with varying base learner and backbone architecture. The former group of results used the standard 4-layer convolutional network with 64 filters per layer used in \cite{matching-net, proto-net}, whereas the latter used a 12-layer ResNet without the global average pooling. ``RR" stands for ridge regression.}
\label{tab:imagenet-tradeoff}
\begin{center}
\begin{small}
\renewcommand{\arraystretch}{1.3}\setlength{\tabcolsep}{2pt}\begin{tabular}{@{}l@{}c@{}cccc@{}c@{}cccc@{}}
\hline
\toprule
& \phantom{ab} & \multicolumn{4}{c}{\textbf{miniImageNet 5-way}} &  \phantom{ab} & \multicolumn{4}{c}{\textbf{tieredImageNet 5-way}}\\
\cmidrule{3-6} \cmidrule{8-11}
& & \multicolumn{2}{c}{\textbf{1-shot}} & \multicolumn{2}{c}{\textbf{5-shot}} & & \multicolumn{2}{c}{\textbf{1-shot}} & \multicolumn{2}{c}{\textbf{5-shot}} \\
\textbf{model} & & \textbf{acc. (\%)} & \textbf{time (ms)} & \textbf{acc. (\%)} & \textbf{time (ms)} & & \textbf{acc. (\%)} & \textbf{time (ms)}  & \textbf{acc. (\%)} & \textbf{time (ms)}  \\

\hline
\multicolumn{7}{@{}l}{\textbf{4-layer conv (feature dimension=1600)} } \\
  \quad Prototypical Networks \cite{Mensink:2013:DIC:2554063.2554069, proto-net}  & & 53.47$_{\pm \text{0.63}}$ & 6$_{\pm \text{0.01}}$ & 70.68$_{\pm \text{0.49}}$ & 7$_{\pm \text{0.02}}$ & & 54.28$_{\pm \text{0.67}}$  & 6$_{\pm \text{0.03}}$  & 71.42$_{\pm \text{0.61}}$  & 7$_{\pm \text{0.02}}$  \\
  \quad MetaOptNet-RR (ours) & & 53.23$_{\pm \text{0.59}}$  & 20$_{\pm \text{0.03}}$ & 69.51$_{\pm \text{0.48}}$ & 27$_{\pm \text{0.05}}$ & & 54.63$_{\pm \text{0.67}}$ & 21$_{\pm \text{0.05}}$  & 72.11$_{\pm \text{0.59}}$ & 28$_{\pm \text{0.06}}$ \\
  \quad MetaOptNet-SVM (ours) & & 52.87$_{\pm \text{0.57}}$  & 28$_{\pm \text{0.02}}$ & 68.76$_{\pm \text{0.48}}$ & 37$_{\pm \text{0.05}}$ & & 54.71$_{\pm \text{0.67}}$ & 28$_{\pm \text{0.07}}$ & 71.79$_{\pm \text{0.59}}$ & 38$_{\pm \text{0.08}}$ \\
\hline
\multicolumn{7}{@{}l}{\textbf{ResNet-12 (feature dimension=16000)} } \\
  \quad Prototypical Networks \cite{Mensink:2013:DIC:2554063.2554069, proto-net}  & & 59.25$_{\pm \text{0.64}}$  & 60$_{\pm \text{17}}$ & 75.60$_{\pm \text{0.48}}$  & 66$_{\pm \text{17}}$ & & 61.74$_{\pm \text{0.77}}$ & 61$_{\pm \text{17}}$ & 80.00$_{\pm \text{0.55}}$ & 66$_{\pm \text{18}}$  \\
  \quad MetaOptNet-RR (ours) & & 61.41$_{\pm \text{0.61}}$  & 68$_{\pm \text{17}}$  & \textbf{77.88}$_{\pm \text{0.46}}$  & 75$_{\pm \text{17}}$  & & \textbf{65.36}$_{\pm \text{0.71}}$ & 69$_{\pm \text{17}}$ & \textbf{81.34}$_{\pm \text{0.52}}$ & 77$_{\pm \text{17}}$ \\
  \quad MetaOptNet-SVM (ours) & & \textbf{62.64}$_{\pm \text{0.61}}$ & 78$_{\pm \text{17}}$ & \textbf{78.63}$_{\pm \text{0.46}}$ & 89$_{\pm \text{17}}$ & & \textbf{65.99}$_{\pm \text{0.72}}$ & 78$_{\pm \text{17}}$ & \textbf{81.56}$_{\pm \text{0.53}}$ & 90$_{\pm \text{17}}$\\
  \bottomrule
\hline
\end{tabular}
\end{small}
\end{center}
\end{table*}

\begin{figure}
\begin{center}
\begin{tabular}{cc}
\hspace{-6mm}\includegraphics[height=0.18\textwidth]{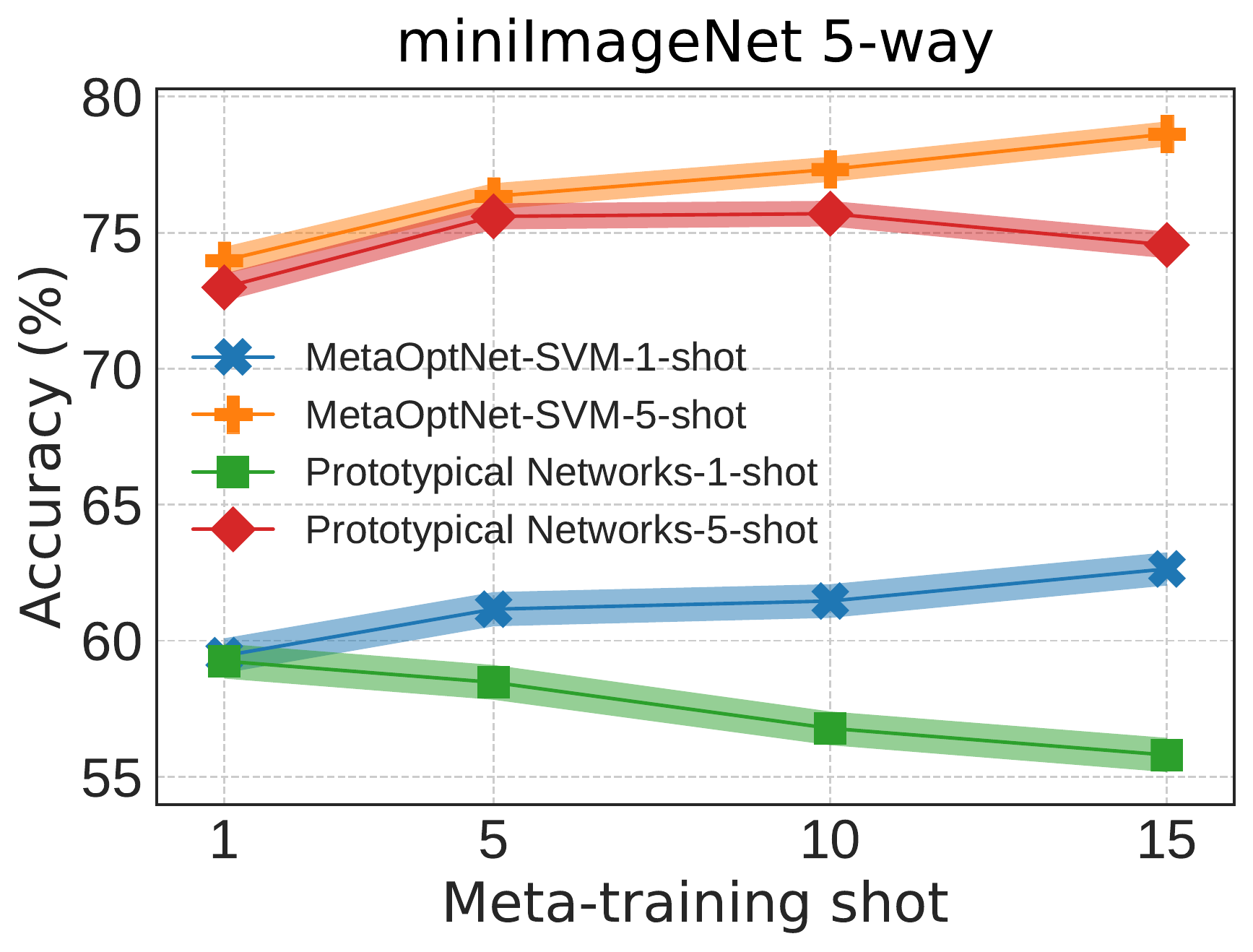} & \hspace{-3.5mm}
\includegraphics[height=0.18\textwidth]{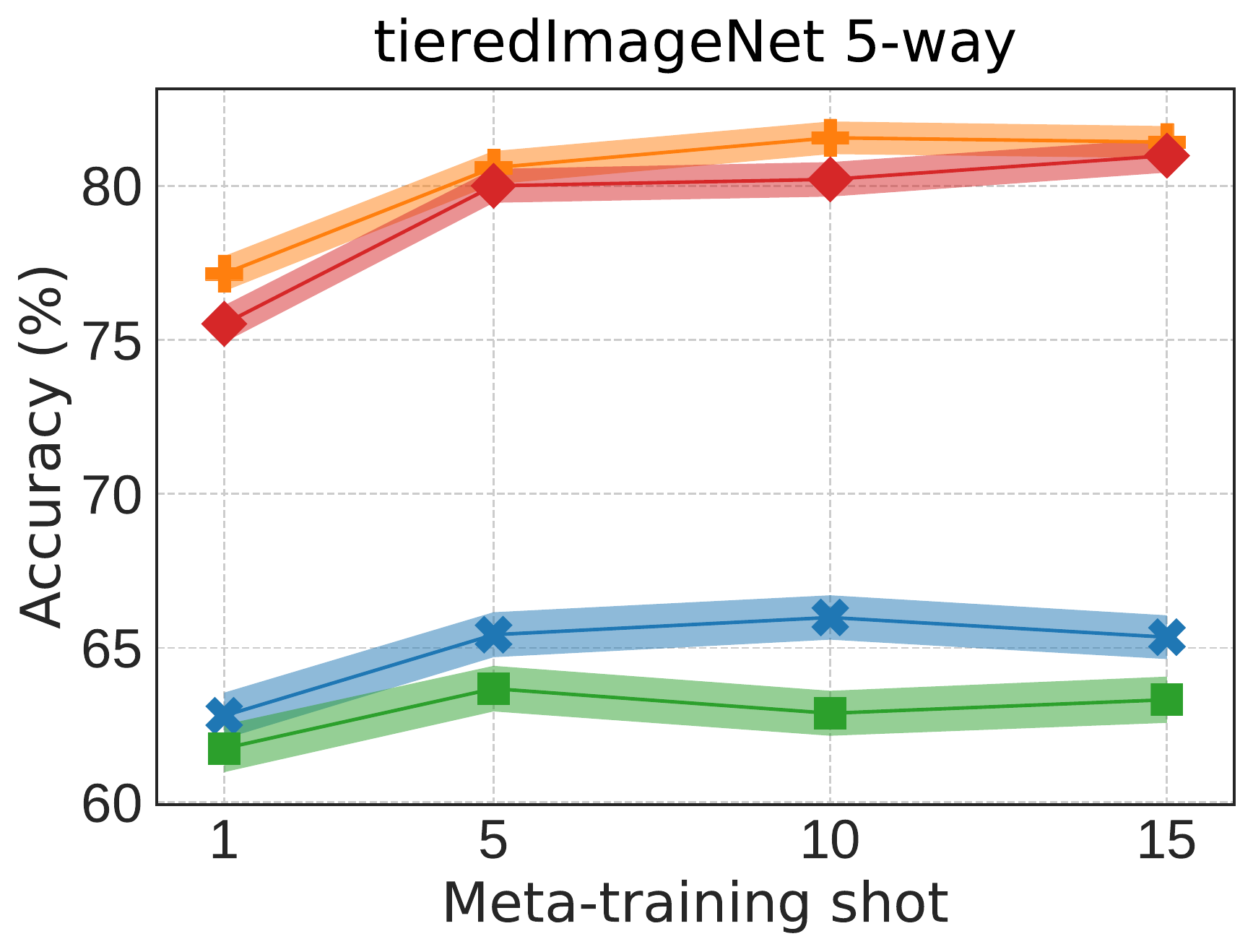}  \hspace{-7mm} \\
\hspace{-6mm}\includegraphics[height=0.18\textwidth]{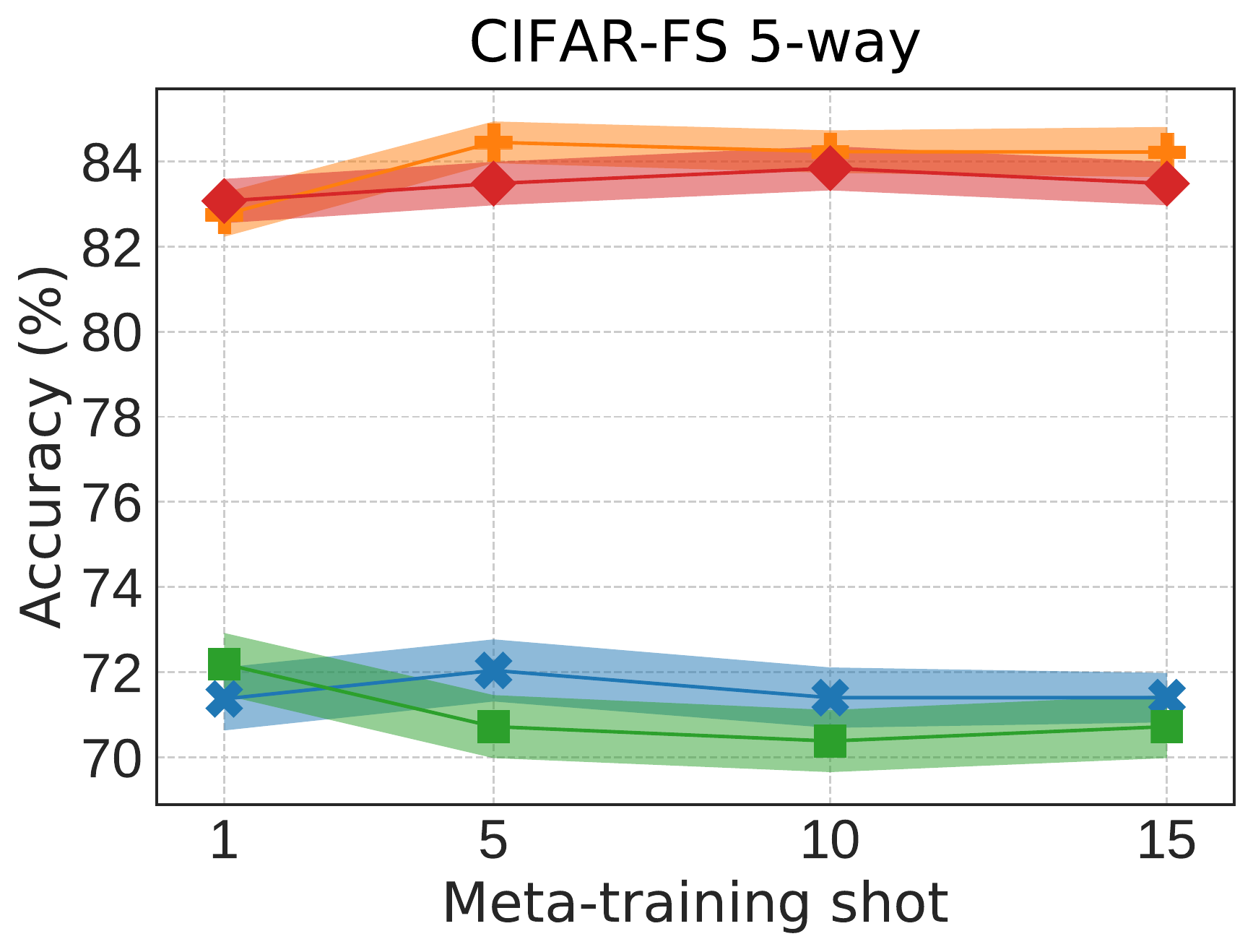} & \hspace{-5mm}
\includegraphics[height=0.18\textwidth]{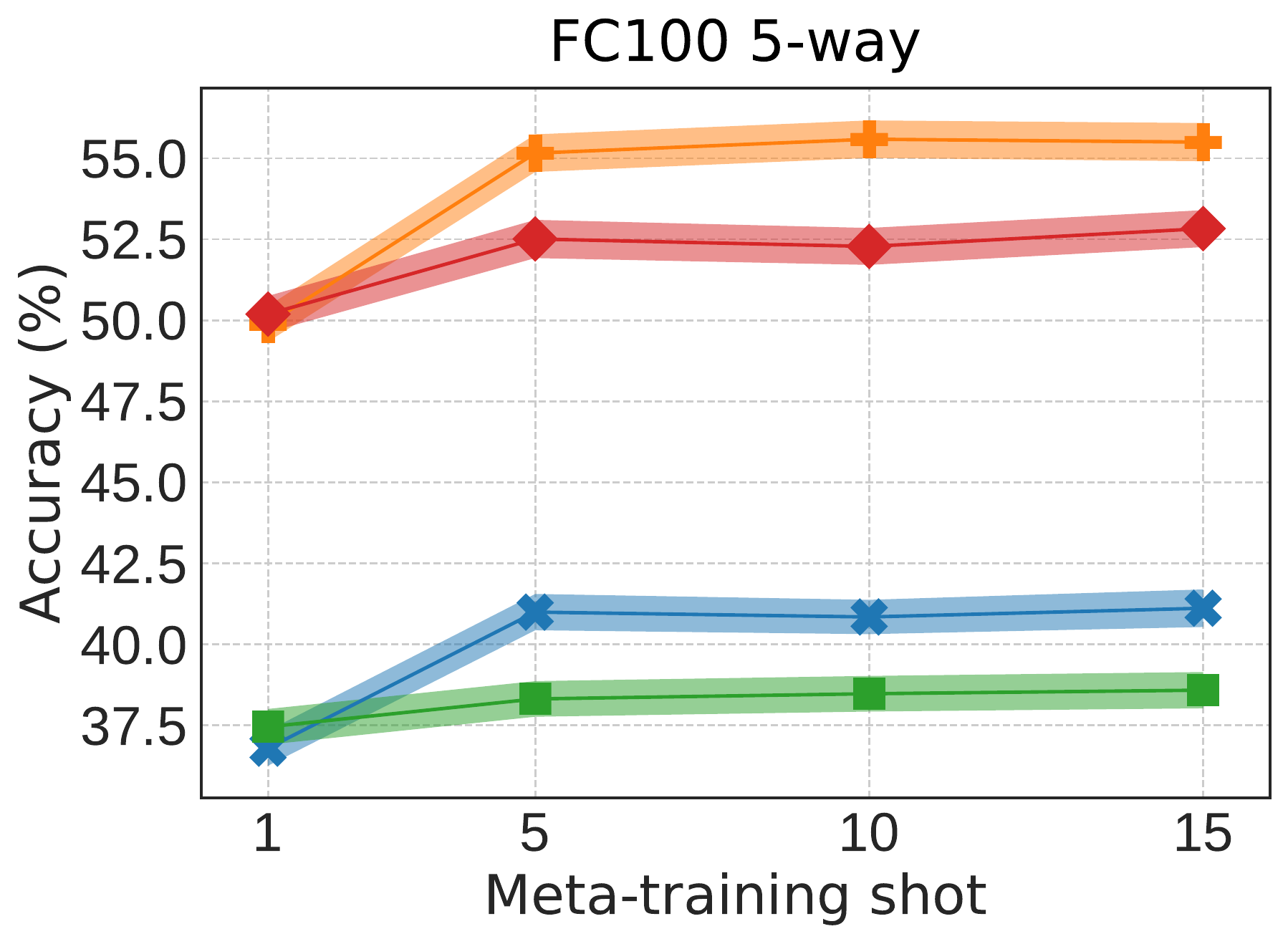}  \hspace{-7mm}
\end{tabular}
\caption{\textbf{Test accuracies (\%) on meta-test
    sets with varying meta-training shot}. Shaded region denotes 95\%
  confidence interval. In general, the performance of MetaOptNet-SVM on both
1-shot and 5-shot regimes increases with increasing
meta-training shot.}
\label{fig:meta-training-shot}
\vspace{-7mm}
\end{center}
\end{figure}

\subsection{Comparisons between base learners}\label{sec:exp:ablation}
Table \ref{tab:imagenet-tradeoff} shows the
results where we vary the base learner for two different
embedding architectures.
When we use a standard 4-layer convolutional network where
the feature dimension is low (1600), we do not observe a substantial benefit of
adopting discriminative classifiers for few-shot learning. 
Indeed, nearest class mean classifier \cite{Mensink:2013:DIC:2554063.2554069}
is proven to work well under a low-dimensional feature as shown in
Prototypical Networks \cite{proto-net}. 
However, when the embedding dimensional is much higher (16000), SVMs yield
better few-shot accuracy than other base learners.
Thus, regularized linear classifiers provide robustness when
high-dimensional features are available. 

The added benefits come at a modest increase in computational cost.
For ResNet-12, compared to nearest class mean classifier, the additional
overhead is around 13\% for the ridge regression base
learner and around 30-50\% for the SVM base learner. 
As seen in from Figure \ref{fig:meta-training-shot}, the performance of our model on both
1-shot and 5-shot regimes generally increases with increasing
meta-training shot. This makes the approach more practical as we can
meta-train the embedding once with a high shot for all meta-testing shots.

As noted in the FC100 experiment, SVM base learner seems to be
beneficial when the semantic overlap between test and train is
smaller. 
We hypothesize that the class embeddings are more significantly more compact for
training data than test data (\eg, see ~\cite{Yosinski:2014:TFD:2969033.2969197});
hence flexibility in the base learner allows robustness to noisy
embeddings and improves generalization.

\subsection{Reducing meta-overfitting}\label{sec:exp:overfitting}
\noindent\textbf{Augmenting meta-training set.} 
Despite sampling tasks, at the end of meta-training MetaOptNet-SVM with
ResNet-12 achieves nearly $100$\% test accuracy on all the meta-training datasets except the tieredImageNet. 
To alleviate overfitting, similarly to
\cite{LEO,Act2Param}, we use the union of the meta-training and
meta-validation sets to meta-train the embedding, keeping the
hyperparameters, such as the number of epochs, identical to the previous setting.
In particular, we terminate the meta-training
after 21 epochs for miniImageNet, 52 epochs for tieredImageNet, 21
epochs for CIFAR-FS, and 21 epochs for FC100. 
Tables~\ref{tab:miniImagenet} and \ref{tab:CIFAR} show the results with
the augmented meta-training sets, denoted as MetaOptNet-SVM-trainval. On
minImageNet, CIFAR-FS, and FC100 datasets, we observe improvements in
test accuracies. On tieredImageNet dataset, the difference is
negligible. 
We suspect that this is because our system has not yet
entered the regime of overfitting (In fact, we observe $\sim$94$\%$
test accuracy on tieredImageNet meta-training set). Our results
suggest that meta-learning embedding with more meta-training
\emph{``classes"} helps reduce overfitting to the meta-training
set.

\noindent\textbf{Various regularization techniques.} Table
\ref{table:ablation} shows the effect of regularization methods on
MetaOptNet-SVM with ResNet-12. 
We note that early works on few-shot
learning \cite{proto-net, MAML} did not employ any of these
techniques. We observe that without the use of regularization, the
performance of ResNet-12 reduces to the one of the 4-layer
convolutional network with 64 filters per layer shown in Table
\ref{tab:imagenet-tradeoff}. This shows the importance of regularization
for meta-learners. We expect that performances of few-shot learning
systems would be further improved by introducing novel regularization
methods.

\begin{figure}
\begin{center}
\begin{tabular}{cc}
\hspace{-8mm}\includegraphics[height=0.173\textwidth]{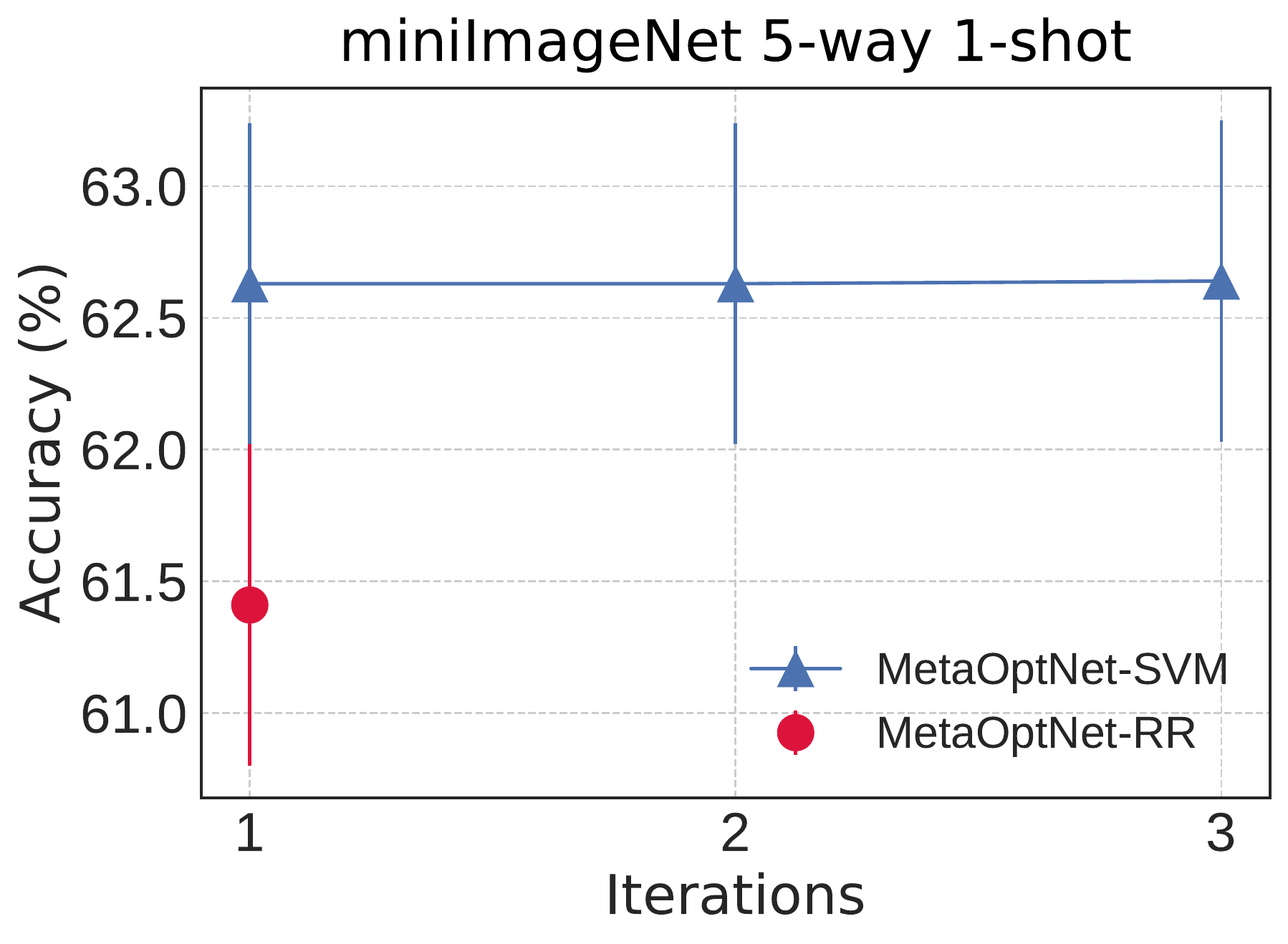} & \hspace{-5mm}
\includegraphics[height=0.173\textwidth]{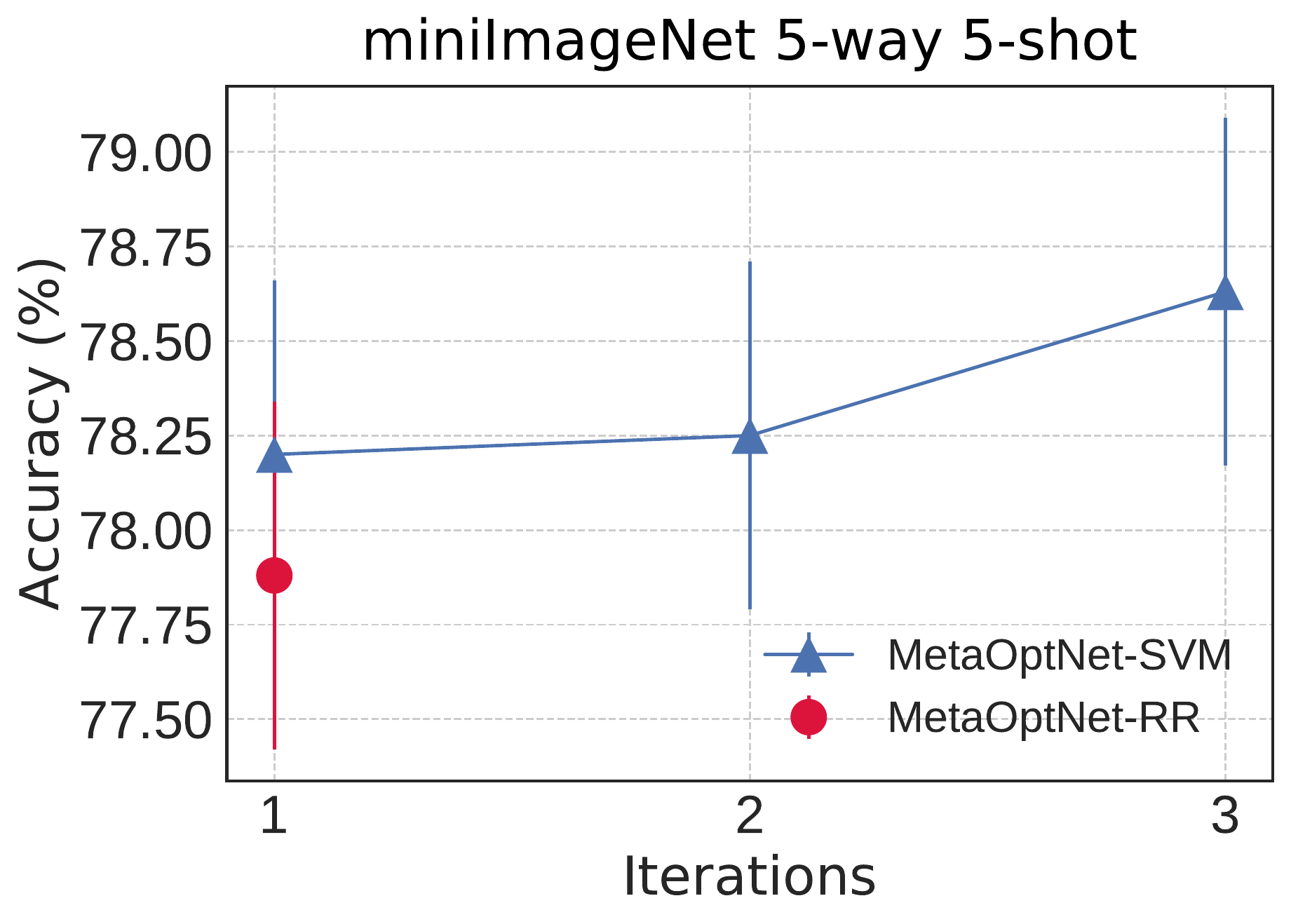}  \hspace{-10mm}
\end{tabular}
\caption{\textbf{Test accuracies (\%) on miniImageNet meta-test set with varying iterations of QP solver.} The error bar denotes 95\% confidence interval. Ridge regression base learner (MetaOptNet-RR)
  converges in 1 iteration; SVM base learner (MetaOptNet-SVM)
  was run for 3 iterations.}
\label{fig:learning-curve}

\end{center}
\vspace{-7mm}
\end{figure}

\subsection{Efficiency of dual optimization}\label{sec:exp:efficiency}
To see whether the dual optimization is indeed effective and
efficient, we measure accuracies on meta-test set with varying
iteration of the QP solver. Each iteration of QP solver
\cite{amos2017optnet} involves computing updates for primal and dual
variables via LU decomposition of KKT matrix. 
The results are shown in
Figure \ref{fig:learning-curve}. 
The QP solver reaches the optima of ridge regression objective in
just one iteration. Alternatively one can use its closed-form
solution as used in~\cite{R2D2}. 
Also, we observe that for 1-shot tasks, the QP SVM solver reaches
optimal accuracies in 1 iteration, although we observed that the KKT
conditions are not exactly satisfied yet. 
For 5-shot tasks, even if we run QP SVM solver for 1 iteration, we achieve better accuracies than other
base learners. When the iteration of SVM solver is limited to 1
iteration, 1 episode takes 69 $\pm$ 17 ms for an 1-shot task, and 80
$\pm$ 17 ms for a 5-shot task, which is on par with the computational
cost of the ridge regression solver (Table
\ref{tab:imagenet-tradeoff}). These experiments show that solving dual
objectives for SVM and ridge regression is very effective under
few-shot settings.

\section{Conclusion}
In this paper, we presented a meta-learning approach with convex base learners for few-shot learning.
The dual formulation and KKT conditions can be exploited to
enable computational and memory efficient meta-learning that is
especially well-suited for few-shot learning problems.
Linear classifiers offer better generalization than nearest-neighbor
classifiers at a modest increase in computational costs (as seen in
Table~\ref{tab:imagenet-tradeoff}).
Our experiments suggest that regularized linear models allow
significantly higher embedding dimensions with reduced
overfitting.
For future work, we aim to explore other convex base-learners such as
kernel SVMs. 
This would allow the ability to incrementally increase model capacity as
more training data becomes available for a task.

\begin{table}
\centering
\resizebox{\linewidth}{!}
{%
\setlength{\tabcolsep}{2.5pt}
\begin{tabular}{@{}ccccc|cc@{}}

\hline
\toprule
\begin{tabular}{c} \textbf{Data} \\ \textbf{Aug.}\end{tabular} & \begin{tabular}{c} \textbf{Weight} \\ \textbf{Decay} \end{tabular} & \begin{tabular}{c} \textbf{Drop} \\ \textbf{Block} \end{tabular} & \begin{tabular}{c} \textbf{Label} \\ \textbf{Smt.}\end{tabular}& \begin{tabular}{c} \textbf{Larger} \\ \textbf{Data} \end{tabular} &  \textbf{1-shot} & \textbf{5-shot} \\
\midrule
&  &  &  & & 51.13 & 70.88 \\
\checkmark &  &  &  & & 55.80 & 75.76 \\
 & \checkmark &  &  & & 56.65 & 73.72 \\
\checkmark & \checkmark &  &  & & 60.33 & 76.61 \\
\checkmark & \checkmark & \checkmark &  & & 61.11 & 77.40 \\
\checkmark & \checkmark & \checkmark & \checkmark &   & 62.64 & 78.63 \\
\checkmark & \checkmark & \checkmark & \checkmark & \checkmark  & 64.09 & 80.00 \\

\bottomrule
\hline
\end{tabular}
}
\caption{{\textbf{Ablation study.} Various regularization techniques improves test accuracy regularization techniques} improves test accuracy (\%) on 5-way miniImageNet benchmark. We use MetaOptNet-SVM with ResNet-12 for results. `Data Aug.', `Label Smt.', and `Larger Data' stand for data augmentation, label smoothing on the meta-learning objective, and merged dataset of meta-training split and meta-test split, respectively.}
\vspace{-5mm}
\label{table:ablation}
\end{table}

\noindent {\bf Acknowledgements}. The authors thank Yifan Xu, Jimmy
Yan, Weijian Xu, Justin Lazarow, and Vijay Mahadevan for valuable discussions. Also,
we appreciate the anonymous reviewers for their helpful and
constructive comments and suggestions. Finally, we would like to thank
Chuyi Sun for help with Figure~\ref{fig:MetaOptNet_pipeline}.

{\small
\bibliography{metaoptnet}
\bibliographystyle{ieee_fullname}
}

\end{document}